\pdfoutput=1
\documentclass[11pt]{article}

\usepackage{setspace}
\usepackage{emnlp2021}
\usepackage{times}
\usepackage{latexsym}

\usepackage[T1]{fontenc}
\usepackage[utf8]{inputenc}

\usepackage{microtype}

\usepackage{color}

\usepackage{amsmath}
\usepackage{booktabs}
\usepackage[inline]{enumitem}
\usepackage{graphicx}
\usepackage{inconsolata}
\usepackage{makecell}
\usepackage{multirow}
\usepackage{todonotes}
\usepackage{xcolor}

\usepackage{longtable}
\usepackage{supertabular}

\graphicspath{{img/},}

\title{Cross-Attention is All You Need:\\Adapting Pretrained Transformers for Machine Translation}
\author{Mozhdeh Gheini, Xiang Ren, Jonathan May \\
  Information Sciences Institute \\
  University of Southern California \\
  \texttt{\{gheini, xiangren, jonmay\}@isi.edu}}
\begin{document}
\maketitle
\begin{abstract}
\vspace{-0.1cm}
% to be refined as I go: aligned embeddings mention? mitigate forgetting? zero-shot?
% ask Jon about his comment here
% We study the power of \textit{cross-attention} in the Transformer architecture within the context of machine translation. In transfer learning experiments, where we fine-tune a translation model on a new language pair with either a new source or target language, we find that, apart from the new language's embeddings, only the cross-attention parameters need to be fine-tuned to obtain competitive BLEU performance. We provide insights
We study the power of \textit{cross-attention} in the Transformer architecture within the context of \textit{transfer learning} for machine translation, and extend the findings of studies into cross-attention when \textit{training from scratch}.
% \xiang{follow with a sentence about the motivation (why this matters/what draws you to look at this)}
We conduct a series of experiments through fine-tuning a translation model on data where either the source or target language has changed. These experiments reveal that fine-tuning only the cross-attention parameters is nearly as effective as fine-tuning all parameters (i.e., the entire translation model). We provide insights into why this is the case and observe that limiting fine-tuning in this manner yields cross-lingually aligned embeddings. The implications of this finding for researchers and practitioners include a mitigation of catastrophic forgetting, the potential for zero-shot translation, and the ability to extend machine translation models to several new language pairs with reduced parameter storage overhead.\footnote{Our code is available at \url{https://github.com/MGheini/xattn-transfer-for-mt}.}
% \xiang{say a bit more about 1) the findings of why this is the case; and 2) the implications of the findings (to researchers and to practitioners).}
\end{abstract}

\section{Introduction}
\label{sec:intro}

\vspace{-0.1cm}
The Transformer \cite{NIPS2017_3f5ee243} has become the de facto architecture to use across tasks with sequential data. It has been dominantly used for natural language tasks, and has more recently also pushed the state-of-the-art on vision tasks \cite{dosovitskiy2021an}. In particular, transfer learning from large pretrained Transformer-based language models has been widely adopted to train new models: adapting models such as BERT \cite{devlin-etal-2019-bert} and XLM-R \cite{conneau-etal-2020-unsupervised} for encoder-only tasks and models such as BART \cite{lewis-etal-2020-bart} and mBART \cite{liu-etal-2020-multilingual} for encoder-decoder tasks like machine translation (MT). This transfer learning is predominantly performed in the form of fine-tuning: using the values of several hundred million parameters from the pretrained model to initialize a model and start training from there.

% \begin{figure}[t]
% \centering
%   \includegraphics[width=\columnwidth]{4trans.png}
% \caption{Overview of our transfer learning experiments, depicting (a) training from \texttt{scratch}, (b) conventional fine-tuning (\texttt{src+body}), (c) fine-tuning cross-attention (\texttt{src+xattn}), (d) fine-tuning new vocabulary (\texttt{src}). Dotted components are initialized randomly, while solid lines are initialized with parent parameters. Shaded, underlined components are fine-tuned, while other components are frozen.} 
% %{Transformer modules. During transfer learning, the model is fine-tuned on a new dataset, with all parameters being treated the same way. The embeddings, the encoder and decoder bodies, and the cross-attention all get updated. In this work we consider several fine-tuning strategies that include or exclude modules at the granularity shown above.}
% \label{fig:motive}
% \end{figure}

\begin{figure*}[t]
\centering
  \includegraphics[width=\textwidth]{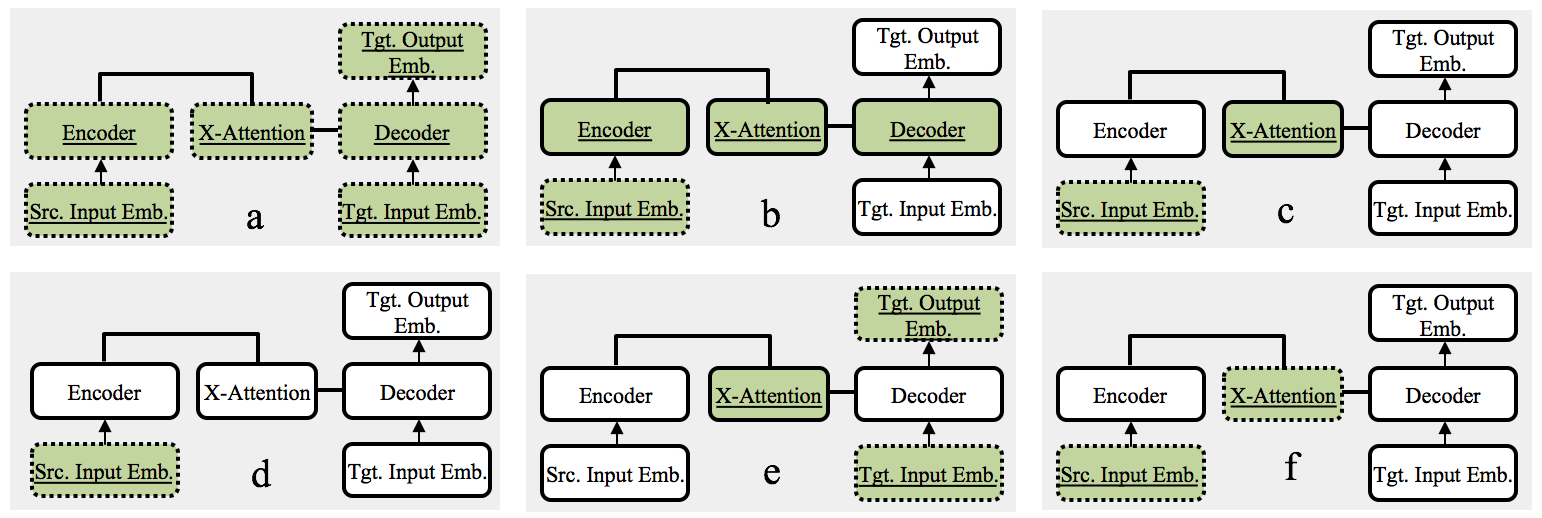}
\caption{Overview of our transfer learning experiments, depicting (a) training from \texttt{scratch}, (b) conventional fine-tuning (\texttt{src+body}), (c) fine-tuning cross-attention (\texttt{src+xattn}), (d) fine-tuning new vocabulary (\texttt{src}), (e) fine-tuning cross-attention when transferring target language (\texttt{tgt+xattn}), (f) transfer learning with updating cross-attention from scratch (\texttt{src+randxattn}). Dotted components are initialized randomly, while solid lines are initialized with parameters from a pretrained model. Shaded, underlined components are fine-tuned, while other components are frozen.} 
%{Transformer modules. During transfer learning, the model is fine-tuned on a new dataset, with all parameters being treated the same way. The embeddings, the encoder and decoder bodies, and the cross-attention all get updated. In this work we consider several fine-tuning strategies that include or exclude modules at the granularity shown above.}
\label{fig:motive}
\end{figure*}

% \xiang{transition into this paragraph reads odd. You just motivated the importance of cross-attn for MT in last paragraph; may be here you should talk about the gap left in the SOTA MT systems to motivate your study?}
Fine-tuning pretrained models often involves updating all parameters of the model without making a distinction between them based on their importance. However, copious recent studies have looked into the relative cruciality of multi-headed self- and cross- attention layers when training an MT model \textit{from scratch} \cite{voita-etal-2019-analyzing,NEURIPS2019_2c601ad9,you-etal-2020-hard}.
% These studies find that the existence of self-attention layers is required because of their role in evolving contextualized representations \cite{tang-etal-2019-understanding}. However,
Cross-attention (also known as encoder-decoder attention) layers are more \textit{important} than self-attention layers in the sense that they result in more degradation in quality when pruned, and hence, are more sensitive to pruning \cite{voita-etal-2019-analyzing,NEURIPS2019_2c601ad9}. Also, cross-attention cannot be replaced with hard-coded counterparts (e.g., an input-independent Gaussian distribution) without significantly hurting the performance, while self-attention can \cite{you-etal-2020-hard}. With the ubiquity of fine-tuning as a training tool, we find a similar investigation focused on transfer learning missing. In this work, we inspect cross-attention and its importance and capabilities through the lens of transfer learning for MT.

% \xiang{I feel you can start a 3rd paragraph for introducing specifics about your study, including how you break it down to multiple analysis, the detailed setups, and so.}
At a high level, we look at training a model for a new language pair by transferring from a pretrained MT model built on a different language pair.
% (in our experiments this difference originates from changing the source or target language, but not both)
 Given that, our study frames and addresses three questions:
% \xiang{can collapse to inline.}
\begin{enumerate*}[label=\arabic*)]
    \item How \textit{powerful} is cross-attention alone in terms of adapting to the new language pair while other modules are frozen?
    \item How \textit{crucial} are the cross-attention layers pretrained values with regard to successful adaptation to the new task? and
    \item Are there any \textit{qualitative differences} in the learned representations when cross-attention is the only module that gets updated?
\end{enumerate*}

To answer these questions, we compare multiple strategies of fine-tuning towards a new language pair from a pretrained translation model that shares one language with the new pair. These are depicted in Figure~\ref{fig:motive}:
% \xiang{can collapse to inline.}
\begin{enumerate*}[label=\alph*)]
    \item Ignoring the pretrained parameters and training entirely from randomly initialized parameters (i.e. `from scratch')
    \item Fine-tuning all parameters except the embeddings for the language in common,\footnote{Freezing shared language embeddings is common practice \cite{zoph-etal-2016-transfer}.} (i.e. `regular' fine-tuning, our upper bound),
    \item Fine-tuning solely the cross-attention layers and new embeddings, and
    \item Fine-tuning only the new embeddings.
\end{enumerate*}
Here, new embeddings refer to randomly initialized embeddings corresponding to the vocabulary of the new language. In Figures~\ref{fig:motive}a--\ref{fig:motive}d, we assume the new language pair has a new source language and not a new target language; Figure~\ref{fig:motive}e shows an example of target-side transfer. In the experiments that follow we will always train new, randomly initialized embeddings for the vocabulary of the newly introduced language. Generally, all other parameters are imported from a previously built translation model and, depending on the experiment, some will remain unchanged and others will be adjusted during training. 
% More specifically, when transferring a pretrained model (in our case, on Fr-En) to another language pair (e.g., Ro-En), we first re-initialize the new source (or target) embeddings from scratch to account for the new source (or target) language. We then initialize the rest of the Transformer model with the pretrained model parameters.

% maybe provide quantification here
Our experiments and analyses show that %in addition to updating randomly re-initialized embeddings, 
fine-tuning the cross-attention layers while keeping the encoder and decoder fixed results in MT quality that is close to what can be obtained when fine-tuning all parameters (\S\ref{sec:exp}).
% This suggests that the cross-attention layers are effectively capable of taking advantage of directly-transferable translation knowledge encoded in the pretrained model and adapting towards the new task.
Evidence also suggests that fine-tuning the \textit{previously trained} cross-attention values is in fact important---if we start with randomly initialized cross-attention parameter values instead of the pretrained ones, we see a quality drop.

Furthermore, intrinsic analysis of the embeddings learned under the two scenarios reveals that full fine-tuning exhibits different behavior from cross-attention-only fine-tuning. When the encoder and decoder bodies are not fine-tuned, we show that the new language's newly-learned embeddings \textit{align} with the corresponding embeddings in the pretrained model. That is, when we transfer from Fr--En to Ro--En for instance, the resulting Romanian embeddings are aligned with the French embeddings. However, we do not observe the same effect when fine-tuning the entire body. In \S\ref{sec:disc} we see how such aligned embeddings can be useful. We specifically show they can be used to alleviate forgetting and perform zero-shot translation.

Finally, from a practical standpoint, our strategy of fine-tuning only cross-attention is also a more \textit{lightweight} fine-tuning approach \cite{houlsby2019parameter} that reduces the storage overhead for extending models to new language pairs: by fine-tuning a subset of parameters, we only need to keep a copy of those instead of a whole-model's worth of values for the new pair. We quantify this by reporting the fraction of parameters that is needed in our case relative to having to store a full new model for each adapted task.

Our \textbf{contributions} are: 1) We empirically show the competitive performance of exclusively fine-tuning the cross-attention layers when contrasted with fine-tuning the entire Transformer body; 
2) We show that when fine-tuning only the cross-attention layers, the new embeddings get aligned with the respective embeddings in the pretrained model. The same effect does not hold when fine-tuning the entire Transformer body;
3) we demonstrate effective application of this aligning artifact in mitigating catastrophic forgetting \cite{Goodfellow14anempirical} and zero-shot translation.

% % \xiang{can collapse to inline.}
% \begin{itemize}
%     \item We empirically show the competitive performance of exclusively fine-tuning cross-attention layers when contrasted with fine-tuning the entire Transformer body.
%     \item We show that when fine-tuning only cross-attention layers, the new embeddings get aligned with the respective embeddings in the pretrained model. The same effect does not hold when fine-tuning the entire Transformer body. % \xiang{what's the implication of this observation? Does alignment bring more improvement?}
%     \item Finally, we investigate two areas where this aligning artifact can be taken advantage of: mitigating catastrophic forgetting \cite{Goodfellow14anempirical} and %the possibility of 
%     zero-shot translation. % \xiang{these two weren't touched in your previous writing}
% \end{itemize}

% \xiang{I recommend to come up a motivating figure on 1st page. Can be illustrating what gap we saw on the current system that motivate us to this study; or anything else you find helpful to give readers a quick idea}

\section{Cross-Attention Fine-Tuning for MT}
\label{sec:expdesign}

% \xiang{Couple comments: 

% 1) One paragraph to re-cap background and goal of the study briefly; 

% 2) 1st subsection (Problem Statement): to more formally define the notations, problem setup, evaluation protocol/metrics. We need some notations to make things clear.

% 3) 2nd subsection (Analysis Setup): several paragraphs will present analysis questions, and corresponding setup of the analysis. We need some notations to make things clear; 

% 4) 3rd \& 4th subsection: datasets \& base models.

% }

Fine-tuning pretrained Transformer models towards downstream tasks has pushed the limits of NLP, and MT has been no exception \cite{liu-etal-2020-multilingual}. Despite the prevalence of using pretrained Transformers, recent studies focus on investigating the importance of self- and cross- attention heads while training models from scratch \cite{voita-etal-2019-analyzing, NEURIPS2019_2c601ad9, you-etal-2020-hard}. These studies verify the relative importance of cross-attention over self-attention heads by exploring either pruning \cite{voita-etal-2019-analyzing, NEURIPS2019_2c601ad9} or hard-coding methods \cite{you-etal-2020-hard}. Considering these results and the popularity of \textit{pretrained} Transformers, our goal in this work is to study the significance of cross-attention while focusing on transfer learning for MT. This section formalizes our problem statement, introduces the notations we will use, and describes our setup to address the questions we raise.

\subsection{Problem Formulation}
\label{subsec:prob}

% \xiang{could you add some paragraph{title} to help segment and navigate the long content?`}

In this work, we focus on investigating the effects of the cross-attention layers when fine-tuning pretrained models towards new MT tasks. Fine-tuning for MT is a transfer learning method that, in its simplest form \cite{zoph-etal-2016-transfer}, involves training a model called the `parent' model on a relatively high-resource language pair, and then using the obtained parameters to initialize a `child model' when further training towards a new, potentially low-resource, language pair. Here, high-resource and low-resource refer to the amount of \textit{parallel} data that is available for the languages. Henceforth, we use `parent' and `child' when referring to training components (e.g., model, data, etc.) in the pretraining and fine-tuning stages, respectively.

\paragraph{Formal Definition.} Consider a model $f_\theta$ trained on the parent dataset, where each training instance $(x_{s_p}, y_{t_p})$ is a pair of source and target sentences in the parent language pair $s_p\text{--}t_p$. Then fine-tuning is the practice of taking the model's parameters $\theta$ from the model $f_\theta$ to initialize another model $g_\theta$. $g_\theta$ is then further optimized on a dataset of $(x_{s_c}, y_{t_c})$ instances in the child language pair $s_c\text{--}t_c$ until it converges to $g_\phi$. We assume either $s_c=s_p$ or $t_c=t_p$ (i.e., child and parent language pairs share one of the source or target sides).

% It is common practice for fine-tuning to further update all parent parameters $\theta$ on the child data without making any distinction between them.
% % Not only does this result in having to store a complete set of new parameters when extending to each new child task but it also hinders looking into modules in terms of their strength and importance in fine-tuning.
% In this work, inspired by studies into cross-attention when training from scratch, we investigate cross-attention when fine-tuning. \cite{voita-etal-2019-analyzing, NEURIPS2019_2c601ad9, you-etal-2020-hard}.
% % Secondly, given that cross-attention layers have fewer parameters than the encoder and decoder layers, they offer potentially more savings in terms of parameter storage space.

\paragraph{Granular Notations.} It is common practice for fine-tuning to further update all parent parameters $\theta$ on the child data without making any distinction between them. We instead consider $\theta$ at a more granular level, namely as:
\begin{center}
    $\theta = \bigcup\{\theta_{\text{src}}, \theta_{\text{tgt}}, \theta_{\text{enc}}, \theta_{\text{dec}}, \theta_{\text{xattn}}\}$
\end{center}

\noindent where $\theta_{\text{src}}$ includes source-language token embeddings, source positional embeddings, and source embeddings layer norm parameters; $\theta_{\text{tgt}}$ similarly includes target-language (tied) input and output token embeddings, target positional embeddings, and target embeddings layer norm parameters; $\theta_{\text{enc}}$ includes self-attention, layer norm, and feed-forward parameters in the encoder stack; $\theta_{\text{dec}}$ includes self-attention, layer norm, and feed-forward parameters in the decoder stack; and $\theta_{\text{xattn}}$ includes cross-attention and corresponding layer norm parameters.
% In the future sections, we use these notations to explain our experiments.

% To streamline the description of experiments we conduct to address the problem and analysis questions in the following sections, here we define some notations to refer to different modules in the transformer architecture:
% \begin{itemize}
%     \item \textbf{\texttt{src}} refers to source embeddings. It collectively includes the source token embeddings, source positional embeddings, and source embeddings layer norm parameters.
%     \item \textbf{\texttt{tgt}} refers to target embeddings. Target input and output embeddings are tied in our models. So this collectively refers to target input token embeddings, target output token embeddings at the end of the top layer of the decoder, target positional embeddings, and target embeddings layer norm parameters.
%     \item \textbf{\texttt{embed}} is a shorthand for \textbf{\texttt{src+tgt}} when we want to refer to both source and target embedding together.
%     \item \textbf{\texttt{body}} refers to encoder layers, decoder layers, and cross-attention layers including all parameters.
%     \item \textbf{\texttt{xattn}} refers only to cross-attention parameters at each layer including all heads.
%     \item \textbf{\texttt{randxattn}} also refers to cross-attention layers. But it signifies that they've been initialized with random values, not parent model values.
% \end{itemize}

% With the problem formalized and our notations explained, we now describe our analysis setup and unroll the research points we address. 

\subsection{Analysis Setup}
\label{subsec:analysis}

% Our two baselines rely on only updating the appropriate embeddings (needed at the very least) and additionally the entire Transformer body. These are referred to as ``\{src,tgt\}'' and ``\{src,tgt\}+body'' following our notations.
Inspections like ours into individual modules of Transformer often rely on introducing some constraints in order to understand the module better. These constraints come in the form of full removal or pruning \cite{tang-etal-2019-understanding,voita-etal-2019-analyzing}, hard-coding \cite{you-etal-2020-hard}, and freezing \cite{DBLP:journals/corr/abs-2010-11859}. We rely on freezing. We proceed by taking pretrained models, freezing certain parts, and recording the effect on performance, measured by BLEU.

Within the framework of our problem, to address the questions raised in \S\ref{sec:intro}, our analysis compares full and partially-frozen fine-tuning for MT under several settings, which we summarize here:

\paragraph{Cross-attention fine-tuning \& embedding fine-tuning comparative performance.} \label{a:s1} This is to realize how much fine-tuning the cross-attention layers helps in addition to fine-tuning respective embeddings alone.

\paragraph{Cross-attention fine-tuning \& full fine-tuning comparative performance.} \label{a:s2} We wish to find out where fine-tuning  cross-attention stands relative to fine-tuning the entire body. This is to confirm whether or not cross-attention alone can adapt to the child language pair while the encoder and decoder layers are frozen.
% We wish to find out how much fine-tuning the cross-attention helps in addition to updating the embeddings alone and where it stands relative to fine-tuning the entire body. This is to confirm whether or not cross-attention alone can adapt to the new language pair while the encoder and decoder layers are frozen. Hence, in addition to the baselines, we also conduct an experiment where the encoder and decoder stacks remain frozen, and only cross-attention layers get updated in addition to respective embeddings. This will be referred to as ``\{src,tgt\}+xattn''. A visual distinction between ``\{src,tgt\}+body'' and ``\{src,tgt\}+xattn'' is provided in Figure~\ref{fig:setups}.

\paragraph{Pretrained cross-attention layers \& random cross-attention layers.} \label{a:s3} We wish to understand how important a role cross-attention's pretrained values play when single-handedly adapting to a new language pair. This determines if the knowledge encoded in cross-attention itself has a part in its power.
% We thus also include an experiment similar to \{src,tgt\}+xattn where, instead of using pretrained values for cross-attention layers, we re-initialize these layers. We refer  to this variant as ``\{src,tgt\}+randxattn''.

\paragraph{Translation cross-attention \& language modelling cross-attention.} \label{a:s4} Finally, we contrast the knowledge encoded in cross-attention learned by different pretraining objectives. This is to evaluate if the knowledge brought about by a different pretraining objective affects the patterns observed from a cross-attention pretrained on MT while fine-tuning for MT.

% \textbf{Practical effectiveness compared against training from scratch.} Finally, practically, it is important for a transferred model to be able to perform better than simply training a model from scratch on the child language pair. That is the whole argument for transfer learning after all. Therefore, each set of experiments also includes a model trained from scratch to show whether each transfer setup is sane or not. These models will be referred to as ``scratch'' from now on.

\section{Experimental Setup}

% \begin{figure}[t]
% \centering
%   \includegraphics[width=\columnwidth]{transfer.png}
% \caption{Transfer setup to be used in our experiments. Boxes with solid lines represent the parameters that use parent model values, and boxes with dashed lines represent the parameters that use random values. Here, we assume the child language pair has a new source language and not a new target language.}
% \label{fig:transfer}
% \end{figure}

% \begin{figure}[t]
% \centering
%   \includegraphics[width=\columnwidth]{xattn_transfer.png}
% \caption{Schematic diagrams of \texttt{src+body} on the top and \texttt{src+xattn} on the bottom. Dashed source embedding modules use random values, and the rest use parent model values. During transfer learning, green blocks (underscored) get updated, and grey blocks (no underscore) remain frozen.
% % Without loss of generality, here we assume source embeddings are re-initialized. When needing to re-initialize target embeddings (i.e., transferring to a pair with a different target language), embedding blocks simply switch places in terms of block color.
% % \xiang{ you can just use one figure and try to deliver the same msg by some color highlight? So then the figure will be a single column}
% }
% \label{fig:setups}
% \end{figure}

In this section, we describe our experiments and the data and model that we use to materialize the analysis outlined in \S\ref{subsec:analysis}.

\subsection{Methods}
\label{subsec:method}

% \xiang{could you add some paragraph{title} to help segment and navigate this part}

We first provide the details of our transfer setup, and then describe the specific fine-tuning baselines and variants used in our experiments.

\paragraph{General Setup.} An important concern when transferring is initializing the embeddings of the new language. When initializing parameters in the child model, there are several ways to address the vocabulary mismatch between the parent and the child model: frequency-based assignment, random assignment \cite{zoph-etal-2016-transfer}, joint (shared) vocabularies \cite{nguyen-chiang-2017-transfer,kocmi-bojar-2018-trivial,neubig-hu-2018-rapid,DBLP:journals/corr/abs-1909-06516,liu-etal-2020-multilingual}, and no assignment at all, which results in training randomly initialized embeddings \cite{aji-etal-2020-neural}. In our experiments, we choose to always use new random initialization for the new embeddings (including token embeddings, positional embeddings, and corresponding layer norm parameters). This decision is made to later let us study what happens to embeddings under each of the settings, independent of any pretraining artifacts that exist in them. For instance, when transferring from Fr--En to \{Ro--En, Fr--Es\}, respectively, all parameters are reused except for \hspace*{0.05mm}\{$\theta_{\text{src}}$, $\theta_{\text{tgt}}\}$,\footnote{We drop the ``respectively'' henceforth and use \{...\} throughout to indicate alternation.} which get re-initialized given the new \hspace*{0.05mm}\{source, target\} language. The side that remains the same (e.g., En when going from Fr--En to Ro--En), uses the parent vocabulary and keeps the corresponding embeddings frozen during fine-tuning.\footnote{Preliminary ablations fine-tuning all embeddings did not change the outcome or conclusions of our experiments.}

\paragraph{Fine-tuning Settings.} With the general transfer setup, we employ different settings in our experiments to address the points in \S\ref{subsec:analysis}. Each fine-tuning method is clarified based on our notations in \S\ref{subsec:prob}
% \xiang{i would frame these as different methods for analysis to get to insights about your analysis questions}
:
\textbf{1)} \texttt{\{src,tgt\}} only updates the embeddings \{$\theta_{\text{src}}$, $\theta_{\text{tgt}}$\} (Figure~\ref{fig:motive}d).
\textbf{2)} \texttt{\{src,tgt\}+body} additionally updates the entire Transformer body (\{$\theta_{\text{src}}$,  $\theta_{\text{tgt}}$\} + $\theta_{\text{enc}}$ + $\theta_{\text{dec}}$ + $\theta_{\text{xattn}}$) (Figure~\ref{fig:motive}b).
\textbf{3)} \texttt{\{src,tgt\}+xattn} only updates the cross-attention layers in addition to the first baseline (\{$\theta_{\text{src}}$, $\theta_{\text{tgt}}$\} + $\theta_{\text{xattn}}$), and keeps the encoder and decoder stacks frozen (Figure~\ref{fig:motive}c, \ref{fig:motive}e). These collectively address the \hyperref[a:s1]{first} and \hyperref[a:s2]{second} settings in \S\ref{subsec:analysis}.
% A visual distinction between \textbf{\texttt{\{src,tgt\}+body}} and \textbf{\texttt{\{src,tgt\}+xattn}} is provided in Figure~\ref{fig:setups}.
\textbf{4)} \texttt{\{src,tgt\}+randxattn} similarly only updates the cross-attention layers in addition to embeddings, but uses randomly initialized values instead of pretrained values (Figure~\ref{fig:motive}f). This addresses the \hyperref[a:s3]{third} setting in \S\ref{subsec:analysis}.
% \begin{itemize}
%     \item The first baseline, \textbf{\texttt{\{src,tgt\}}}, only updates the respective embeddings \{$\theta_{\text{src}}$, $\theta_{\text{tgt}}$\}, which are needed at the very least.
%     \item The second baseline, \textbf{\texttt{\{src,tgt\}+body}}, additionally updates the entire transformer body (\{$\theta_{\text{src}}$,  $\theta_{\text{tgt}}$\} + $\theta_{\text{enc}}$ + $\theta_{\text{dec}}$ + $\theta_{\text{xattn}}$).
%     \item \textbf{\texttt{\{src,tgt\}+xattn}} only updates the cross-attention layers in addition to the first baseline (\{$\theta_{\text{src}}$, $\theta_{\text{tgt}}$\} + $\theta_{\text{xattn}}$), and keeps the encoder and decoder stacks frozen . A visual distinction between \textbf{\texttt{\{src,tgt\}+body}} and \textbf{\texttt{\{src,tgt\}+xattn}} is provided in Figure~\ref{fig:setups}.
%     \item \textbf{\texttt{\{src,tgt\}+randxattn}} similarly only updates the cross-attention layers in addition to respective embeddings, just using randomly initialized values instead of pretrained values.
% \end{itemize}

For all transfer experiments, we also conduct the \texttt{scratch} variant (Figure~\ref{fig:motive} a), where we train a model from scratch on the child dataset. This is to confirm the effectiveness of transfer under each setting.
We conduct all the above experiments using a French--English translation model as parent and transferring to six different child language pairs. In \S\ref{sec:powerimp} we conduct an ablation that substitutes mBART \cite{liu-etal-2020-multilingual} as a parent. mBART is trained with denoising objective in a self-supervised manner. In contrast to a translation model, the cross-attention layers in mBART have thus not been learned using any parallel data. This enables us to distinguish between different pretraining objectives, addressing the \hyperref[a:s4]{fourth} setting in \S\ref{subsec:analysis}.

\subsection{Data and Model Details}

% \xiang{still confused about why Data and Model are mixed in one part? Can we separate into two?}

\begin{table}[t]
    \centering
    \scalebox{0.70}{
    \begin{tabular}[width=\columnwidth]{llll}
        \toprule
         & \makecell[l]{Train Corpus \\ (Sent. Count)} & Test Corpus & Vocab. Size \\
        \hline
        \addlinespace[0.3em]
        \textbf{Ro--En} & \makecell[l]{WMT16 \\ (612.4~K)} & newstest2016 & 16~K / reuse tgt \\
        \addlinespace[0.3em]
        \textbf{Ja--En} & \makecell[l]{IWSLT17 \\ (223.1~K)} & IWSLT17 & 8~K / reuse tgt \\
        \addlinespace[0.3em]
        \textbf{De--En} & \makecell[l]{IWSLT16 \\ (196.9~K)} & IWSLT16 & 8~K / reuse tgt \\
        \addlinespace[0.3em]
        \textbf{Ha--En} & \makecell[l]{ParaCrawl v8 \\ (159.0~K)} & newsdev2021 & 8~K / reuse tgt \\
        \addlinespace[0.3em]
        \textbf{Fr--Es} & \makecell[l]{News Comm. v15 \\ (283.5~K)} & newstest2013 & reuse src / 8~K \\
        \addlinespace[0.3em]
        \textbf{Fr--De} & \makecell[l]{News Comm. v15 \\ (284.1~K)} & newstest2020 & reuse src / 8~K \\
        \bottomrule
    \end{tabular}}
    \caption{Data sources and statistics for each of the child language pairs.} %The sentence count for each training corpus has been rounded to the nearest hundred.}
    \label{tab:data}
\end{table}

\noindent
\textbf{Dataset.}
% \xiang{This whole subsection can be moved to Sec 2, and break down to ``Datasets" and "Base Models" (or sth more suitable).}
For the choice of language pairs and datasets, we mostly follow \newcite{you-etal-2020-hard} (Fr--En, Ro--En, Ja--En, De--En) and additionally include Ha--En, Fr--Es, and Fr--De.
% \xiang{I think we need to make the details self-contained without assuming readers to check that paper or be familiar with that paper. Let's expand on this to contain everything necessary detail.}
We designate Fr--En as the parent language pair and Ro--En, Ja--En, De--En, Ha--En (new source), Fr--Es, Fr--De (new target) as child language pairs. Our Fr--En parent model is trained on the Europarl + Common Crawl subset of WMT14 Fr--En,\footnote{\url{http://statmt.org/wmt14/translation-task.html}} which comprises 5,251,875 sentences. Details and statistics of the data for the child language pairs are provided in Table~\ref{tab:data}.

\paragraph{Model Details.}
% \xiang{model specs, initialization of different model parts.}
We use the Transformer base architecture (6 layers of encoder and decoder with model dimension of 512 and 8 attention heads) for all models,  \cite{NIPS2017_3f5ee243} and the Fairseq \cite{ott2019fairseq} toolkit for all our experiments.

All models rely on BPE subword vocabularies \cite{sennrich2015neural} processed through the SentencePiece \cite{kudo-richardson-2018-sentencepiece} BPE implementation. The vocabulary for the parent model consists of 32K French subwords on the source side, and 32K English subwords on the target side. The sizes of the vocabularies for child models are also reported in Table~\ref{tab:data}. We follow the advice from \newcite{gowda-may-2020-finding} when deciding what vocabulary size to choose, i.e., we choose the maximum number of operations to ensure a minimum of 100 tokens per type.

\begin{table*}[t]
    \centering
    \scalebox{0.9}{
    \begin{tabular}{lrrrrrr}
        \toprule
         & \textbf{Ro--En} & \textbf{Ja--En} & \textbf{De--En} & \textbf{Ha--En} & \textbf{Fr--Es} & \textbf{Fr--De} \\
        % Corpus & WMT16 & IWSLT17 & IWSLT16 & ParaCrawl v8 \\
        % \hline
        % Train / Test Size (Lines) & 612,422/1,999 & 223,108/1,452 & 196,884/993 & 158,968/1000 \\
        % \cmidrule(lr){1-1}
        \cmidrule(lr){2-5}
        \cmidrule(lr){6-7}
        \texttt{scratch} (100\%) & 29.0 & 9.2 & 30.8 & 5.4 & 24.4 & 18.5 \\
        \texttt{\{src,tgt\}} (8\%) & 29.8 & 8.7 & 32.4 & \textbf{8.6} & 21.6 & 11.6 \\
        \texttt{\{src,tgt\}+body} (75\%) & \textbf{31.0} & \textbf{11.8} & \textbf{36.2} & \textbf{8.8} & \textbf{27.3} & \textbf{21.4} \\
        \texttt{\{src,tgt\}+xattn} (17\%) & \textbf{(-0.1) 30.9} & \textbf{(-2.0) 9.8} & \textbf{(-1.2) 35.0} & (-0.4) 8.4 & \textbf{(-0.8) 26.5} & \textbf{(-1.8) 19.6} \\
        \texttt{\{src,tgt\}+randxattn} (17\%) & 27.9 & 8.4 & 33.3 & 7.0 & 26.0 & 18.8 \\
        \bottomrule
    \end{tabular}}
    \caption{BLEU scores for each of the five experiments across six language pairs. Bold numbers indicate the top two scoring approaches. Percentages in parentheses next to fine-tuning strategy is the fraction of parameters that had to be updated and hence stored as new values for future use. Numbers in parentheses next to \texttt{\{src,tgt\}+xattn} scores show the difference from \texttt{\{src,tgt\}+body}.}
    \label{tab:res}
\end{table*}

\section{Results and Analysis}
\label{sec:exp}

Our preliminary empirical results consist of five experiments for each of the child language pairs based on methods described in \S\ref{subsec:method}: \texttt{scratch}, \hspace*{0.05mm}\texttt{\{src,tgt\}}, \texttt{\{src,tgt\}+body}, \hspace*{0.05mm}\texttt{\{src,tgt\}+xattn}, and \texttt{\{src,tgt\}+randxattn}. Our core results, which rely on transferring from the Fr--En parent under each setting, are reported in Table \ref{tab:res}. All scores are detokenized cased BLEU computed using \textsc{Sacre}BLEU \cite{post-2018-call}.\footnote{Signature: BLEU+case.mixed+numrefs.1+smooth.exp \\ +tok.13a+version.1.4.8.}
% Numbers reported for these experiments and inspection of representations learned by each answer the questions in \S\ref{sec:intro}.

\subsection{Cross-attention's \textit{Power} and \textit{Importance}}
\label{sec:powerimp}
% \xiang{this part answers the 1st analysis question (probably the major one) you had; pls name in that way. And merge the Sec 4 into this section because they're the follow-up questions you study.}

\paragraph{Translation Quality.} Table~\ref{tab:res} shows that \hspace*{0.05mm}\texttt{\{src,tgt\}+xattn} substantially improves upon \hspace*{0.05mm}\texttt{\{src,tgt\}} in all but one case (Ha--En), especially when transferring to a pair with a new target language, and is competitive with \texttt{\{src,tgt\}+body} across all six language pairs, suggesting that
% a) there is universal translation knowledge encoded in the Transformer body, and b)
cross-attention is capable of taking advantage of encoded generic translation knowledge in the Transformer body to adapt to each child task. Performance gain from \texttt{\{src,tgt\}} and drop from \texttt{\{src,tgt\}+body} when changing the target language (i.e., Fr--Es and Fr--De) are more pronounced than when transferring the source. This is expected---when changing the target, two out of three cross-attention matrices (key and value matrices) are now exposed to a new language. When transferring source, only the query matrix is exposed to the new language.

\paragraph{Storage.} We also report the fraction of the parameters that need to be updated in each case. This is equivalent to the storage overhead that the training process incurs, as the updated parameters need to be stored to be used later. However, the parameters that are reused are only stored once. The number of parameters updated is dependent on the size of the vocabulary in each experiment, since embeddings for a new vocabulary are included. Hence, the single number reported for each fine-tuning strategy is the average across the six language pairs. \textit{Extending} to new language pairs following \texttt{\{src,tgt\}+xattn} is much more efficient in this regard, as expected. We concretely calculate the number of parameters that need to be stored combined for the six new language pairs: \texttt{\{src,tgt\}+xattn} stores only 124,430,336 parameters compared to \texttt{\{src,tgt\}+body}'s 313,583,616.
% Setting aside the embeddings, \texttt{\{src,tgt\}+xattn} requires 9\% (17 (row 4) - 8 (row 2)) of a whole-model worth of storage to add the capability to translate a new language pair compared to 67\% (75 (row 3) - 8 (row 2)) required by \texttt{\{src,tgt\}+body}.

% Furthermore, it is evident that under neither of the \{src,tgt\}+body  and \{src,tgt\}+xattn settings, the scores drop below training from scratch.

\paragraph{Pretrained and Random Values.} Finally, \texttt{\{src,tgt\}+randxattn} experiments also offer perspective on the importance of translation knowledge encoded in cross-attention itself.
% in it being attuned to getting transferred.
Not only does randomly initialized cross-attention fail to perform as well as pretrained cross-attention when being transferred, but in two cases, it even falls behind training from scratch.

\begin{figure*}[t]
\centering
  \includegraphics[width=\textwidth]{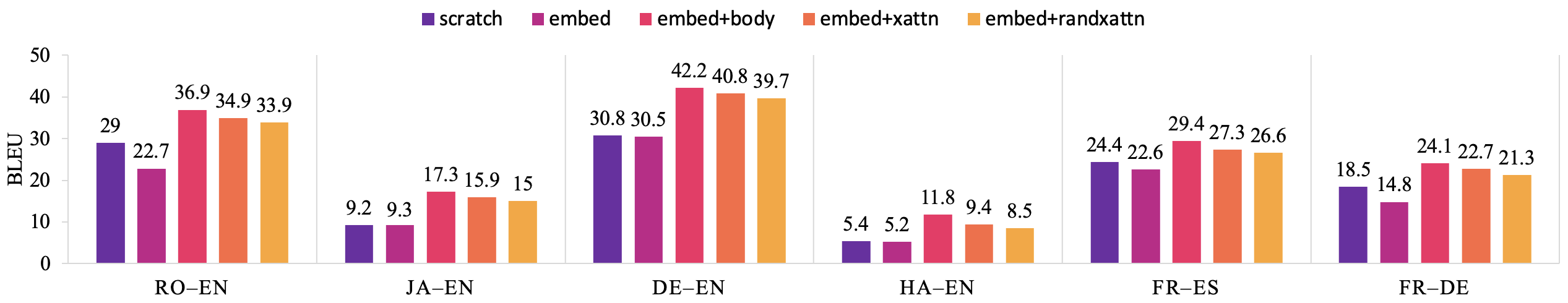}
\caption{BLEU scores across different transfer settings using mBART as parent. Exclusive fine-tuning of embeddings (\texttt{embed}) is not effective at all due to lack of translation knowledge in the cross-attention layers.}
\label{fig:mbart}
\end{figure*}

Our results from transferring mBART \cite{liu-etal-2020-multilingual} to the child language pairs also emphatically illustrate the importance of the type of knowledge encoded in cross-attention. 
% \mozh{is this an appropriate place to discuss mbart experiments in Table~\ref{fig:mbart}? good to include at least another language pair. improve language.}
mBART is a 12-layer Transformer pretrained with a denoising objective in a self-supervised manner using span masking and sentence permutation noising functions. Hence, its cross-attention does not have any \textit{translation} knowledge \textit{a priori}, in contrast with the French--English MT parent model. We transfer mBART to the same language pairs as in Table~\ref{tab:res} and provide the results in Figure~\ref{fig:mbart}. Since mBART uses a shared vocabulary and tied embeddings between the encoder and decoder, in Figure~\ref{fig:mbart} we use \texttt{embed} in experiments' names to signify all embeddings get updated in the case of mBART ($\theta_{\text{src}}$ + $\theta_{\text{tgt}}$).

mBART is a larger model than our Fr--En parent, both in terms of architecture and training data. So a higher range of scores is expected. While the same patterns hold across \texttt{embed+\{body,xattn,randxatnn\}} fine-tuning, the crux of the matter is that \texttt{embed} fine-tuning fails in contrast to the comparable \texttt{\{src, tgt\}} fine-tuning setting of the translation parent. \texttt{src} fine-tuning has higher BLEU than \texttt{scratch} in three cases (Ro--En, De--En, Ha--En). However, \texttt{embed} fine-tuning has higher BLEU than the \texttt{scratch} baseline only in the Ja--En case, and even then, very slightly so (only by 0.1 BLEU). This shows that absence of translation knowledge in mBART's pretrained cross-attention leads to its fine-tuning being more crucial in mBART's functionality for translation adaptation: exclusively fine-tuning embeddings in mBART simply fails, while doing the same with a translation parent model is more successful.
% \mozh{better frame this argument: basically, embed fails miserably cause cross attention is not encoded with translation knowledge. may need to rephrase \{src,tgt\} in Table~\ref{tab:res} as well.}

% \begin{table}[ht]
%     \centering
%     \begin{tabular}[width=\columnwidth]{lc}
%         \toprule
%         \multicolumn{2}{c}{mBART fine-tuned on De-En} \\
%         \midrule
%         mBART\_embed & 30.5 \\
%         mBART\_embed+body & 42.2 \\
%         mBART\_embed+xattn & 40.8 \\
%         mBART\_embed+randxattn & 39.7 \\
%         % mbart\_xattn & \\
%         \bottomrule
%     \end{tabular}
%     \caption{\xiang{create a bar charts for this?}}
%     \label{tab:mbart}
% \end{table}

\subsection{Learned Representations Properties}
\label{subsec:align}

Given that besides cross-attention, embeddings are the only parameters that get updated in both \hspace*{0.05mm}\texttt{\{src,tgt\}+body} and \texttt{\{src,tgt\}+xattn} settings, we take a closer look at them. We want to know how embeddings change under each setting. % We report our findings on the De-En models; however, the reported patterns hold across language pairs.

% We first internally looked at the embedding visualizations. The visualizations implied that under the \{src,tgt\}+xattn setting, new embeddings would clearly align with the corresponding parent embeddings. However, the same was not true under the \{src,tgt\}+body scenario.

To probe the relationship between embeddings learned as a result of different kinds of fine-tuning, we examine the quality of induced\footnote{via nearest neighbor retrieval} bilingual lexicons, a common practice in cross-lingual embeddings literature \cite{artetxe-etal-2017-learning} but incidentally learned in this case. % As our models use  We focus on the subwords that are indeed words; this is easy as we can rely on the whitespace symbol to identify them. We do this to be able to easily assess semantic equivalence against a gold dictionary. For each word in the subword vocabulary, if a word in the gold dictionary, we find its target equivalent based on the nearest neighbor in terms of cosine similarity between the embeddings. We evaluate each retrieved pair by simply checking if it is an exact match with a pair in the gold dictionary or not.

\begin{figure}[ht]
\vspace{-0.2cm}
\centering
  \includegraphics[width=\columnwidth]{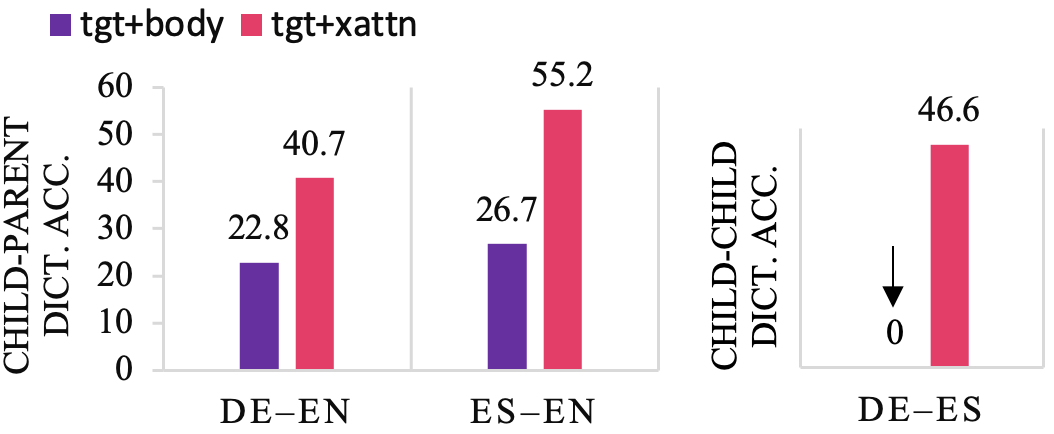}
\caption{Accuracy of bilingual dictionaries induced through embeddings learned under \texttt{tgt+body} and \texttt{tgt+xattn} settings. De and Es effectively get aligned with En under \texttt{tgt+xattn} (left). As they are both aligned to En, we can also indirectly obtain a De--Es dictionary (right). Similar practice completely fails under \texttt{tgt+body}.}
\label{fig:dictacc}
\end{figure}

We use the bilingual dictionaries released as a resource in the MUSE \cite{DBLP:conf/iclr/LampleCRDJ18} repository.\footnote{\url{https://github.com/facebookresearch/MUSE}} For instance, to compare the German embeddings from each of the \texttt{src+body} and \texttt{src+xattn} De--En models to the French embeddings learned in the parent model, we use the De--Fr dictionary. We filter our learned embeddings (which are, in general, of subwords) to be compatible with the MUSE vocabulary. Of the 8,000 German subwords in the vocabulary, 2,025 are found in MUSE. For each of these, we find the closest French embedding by cosine similarity; if the resulting (German, French) pair is in MUSE, we consider this a match. Via this method, we find the accuracy of the bilingual lexicon induction through the embeddings of \texttt{src+xattn} model is 55\%. However, the accuracy through the embeddings of \texttt{src+body} is much lower at 19.7\%. Due to only considering the exact matches against the gold dictionary, this is a very strict evaluation. We also manually look at a sample of 40 words from the German set and check for the correctness of retrieved pairs for those using an automatic translator: while \texttt{src+xattn} scores in the range of ~80\%, \texttt{src+body} scores in the range of ~30\%. Details of this manual inspection are provided in %Table~\ref{tab:mandict}
Table~4 of the appendix. We further report the accuracy of the bilingual dictionaries of three other pairs learned under the two fine-tuning settings for which gold dictionaries are available in Figure~\ref{fig:dictacc}. We don't limit ourselves to child-parent dictionary induction; we also consider child-child dictionary induction (e.g., De--Es) which essentially relies on both languages being aligned with the parent (i.e., En).

Overall, these results confirm that embeddings learned under \texttt{\{src,tgt\}+xattn} effectively get aligned with corresponding parent embeddings. However, this is not the case with embeddings learned under \texttt{\{src,tgt\}+body}. This suggests such effect is not the default pattern in translation models, but rather an artifact of the freezing choices made in \texttt{\{src,tgt\}+xattn}.

\section{Utilities of Aligned Embeddings}
\label{sec:disc}

We saw how fine-tuning only cross-attention results in cross-lingual embeddings with respect to parent embeddings. That is how cross-attention is able to use the baked-in knowledge in the encoder and decoder without any further updates to them. In this section, we discuss two areas where this can be turned to our advantage: mitigating forgetting and performing zero-shot translation.

\subsection{Mitigating Forgetting}

One area where the discovery of \S\ref{subsec:align} can be taken advantage of is mitigating catastrophic forgetting. Catastrophic forgetting refers to the loss of previously acquired knowledge in the model during transfer to a new task. To the best of our knowledge, catastrophic forgetting in MT models has only been studied within the context of inter-domain adaptation \cite{thompson-etal-2019-overcoming,gu-feng-2020-investigating}, and not inter-lingual adaptation.

\begin{figure}[ht]
\vspace{-0.1cm}
\centering
  \includegraphics[width=\columnwidth]{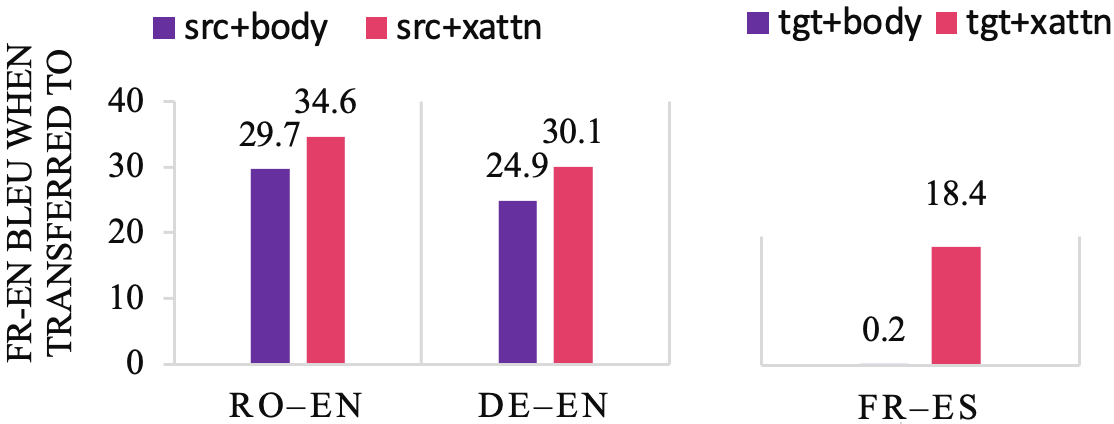}
\caption{Performance on the original language pair after transfer. The original Fr--En parent model scores 35.0 BLEU on the Fr--En test set. \texttt{\{src,tgt\}+xattn} outperforms \texttt{\{src,tgt\}+body} on the parent task.}
\label{fig:forget}
\end{figure}

The effectiveness of the cross-lingual embeddings learned under the  \texttt{\{src,tgt\}+xattn} setting at mitigating forgetting is evident from the results provided in Figure~\ref{fig:forget}. Here we take three of the transferred models, plug back in the appropriate embeddings in them, and compare their performance \textbf{on the original language pair} against the parent model. Specifically, we take the De--En, Ro--En, and Fr--Es models transferred from Fr--En under each of the two \texttt{\{src,tgt\}+xattn} and \texttt{\{src,tgt\}+body settings}, plug in back the original \{Fr, En\} embeddings, and evaluate performance on the Fr--En test set. This score is then compared against the Fr--En parent model performance on Fr--En test set, which scores 35.0 BLEU. While being comparable in terms of performance on the child task as reported in Table~\ref{tab:res}, \texttt{\{src,tgt\}+xattn} constantly outperforms \texttt{\{src,tgt\}+body} on Fr--En. Compared to the original Fr--En model, the source-transferred models (De--En, Ro--En) outperform the target-transferred model (Fr--Es). However, \texttt{tgt+xattn} is much more robust against forgetting compared to \texttt{tgt+body}, which remembers close to nothing (0.2 BLEU).

% \begin{table}[ht]
%     \centering
%     \begin{tabular}[width=\columnwidth]{cccc}
%         \toprule
%          & & \multicolumn{2}{c}{Fr-En BLEU}
%          \\
%          & & \{src,tgt\}+body & \{src,tgt\}+xattn \\
%          \cmidrule(lr){3-3}
%          \cmidrule(lr){4-4}
%          \parbox[t]{5mm}{\multirow{3}{*}{\rotatebox[origin=c]{90}{\makecell{Transfer \\ to}}}} & Ro-En & 29.7 & 34.6 \\
%          & De-En & 24.9 & 30.1 \\
%          & Fr-Es & 0.2 & 18.4 \\
%         \bottomrule
%     \end{tabular}
%     \caption{Performance on the original language pair after transfer under the two studied approaches across three language pairs. The original Fr-En parent model scores 35.0 BLEU on the Fr-En test set. We had established before that the two approaches are comparable in terms of performance on the child language. However,\{src,tgt\}+xattn clearly outperforms \{src,tgt\}+body when it comes to not forgetting. \xiang{would a bar chart be better to deliver the msg?}}
%     \label{tab:forget}
% \end{table}

\subsection{Zero-Shot Translation}

% \xiang{create a small table to include more results?}

Another area where well-aligned embeddings from the \texttt{\{src,tgt\}+xattn} setting can come in handy is zero-shot translation. Since the source embeddings are aligned, we, for instance, can replace the French embeddings in the Fr--Es model learned via \texttt{tgt+xattn} with German embeddings from the De--En model learned via \texttt{src+xattn} and form a De--Es translation model with no De--Es training or direct De--Fr alignment. We additionally build two more zero-shot systems in the same manner: Ro--Es (using transferred Ro--En and Fr--Es models) and Ro--De (using transferred Ro--En and Fr--De models). To put zero-shot scores in context, for each pair we also train a model from scratch: for De--Es using 294,216-sentence News Commentary v14 corpus, and for Ro--Es and Ro--De using 387,653-sentence and 385,663-sentence Europarl corpora respectively. All scores are provided in Table~\ref{tab:zero}.

\begin{table}[ht]
    \centering
    \begin{tabular}[width=0.85\columnwidth]{lrrr}
        \toprule
         & \textbf{De--Es} & \textbf{Ro--Es} & \textbf{Ro--De} \\
        \midrule
        Zero-shot BLEU & 9.2 & 14.7 & 9.8 \\
        Supervised BLEU & 18.3 & 18.6 & 13.4 \\
        \bottomrule
    \end{tabular}
    \caption{Performance of zero-shot systems for three language pairs. De--Es is evaluated on newstest2013 test set. Ro--Es and Ro--De are evaluated on respective TED talks corpus test sets \cite{qi-etal-2018-pre}.}
    \label{tab:zero}
\end{table}

In the case of De--Es, we train two additional models from scratch on 50,000- and 100,000- sentence subsets of the training corpus. These respectively score 7.2 and 12.0 BLEU on the newstest2013 De--Es test set (v.s. zero-shot performance of 9.2). Taken together, these results show that the zero-shot methods we obtain from cross-attention-based transfer can yield reasonable translation models in the absence of parallel data.

\section{Related Work}
% \xiang{can compress to around one column, and move to after Intro.}

% \mozh{need to add a paragraph on lightweight fine-tuning for MT}

% Our work starts off as a study of cross-attention.
\vspace{-0.1cm}
\paragraph{Studying Cross-attention.} Several recent works consider the importance of self- and cross-attention heads in the Transformer architecture \cite{voita-etal-2019-analyzing, NEURIPS2019_2c601ad9, you-etal-2020-hard}. The consensus among these works is that cross-attention heads are relatively more important than self-attention heads when it comes to introducing restrictions in terms of pruning and hard-coding. 
% Specifically, \newcite{voita-etal-2019-analyzing} and \newcite{NEURIPS2019_2c601ad9} concurrently show that pruning a large percentage of encoder self-attention heads minimally affects translation quality. However, cross-attention heads are more sensitive. \newcite{you-etal-2020-hard} also reach a similar conclusion through a different path: instead of pruning, they hard-code attention heads. Their empirical results show that hard-coding self-attention heads does not hurt performance, but hard-coding cross-attention heads leads to significant drops in BLEU.

\paragraph{Module Freezing.} In terms of restrictions introduced, our work is related to a group of recent works that freeze certain modules while fine-tuning \cite{zoph-etal-2016-transfer,artetxe-etal-2020-cross,DBLP:journals/corr/abs-2103-05247}. \newcite{artetxe-etal-2020-cross} conduct their study on an encoder-only architecture. They show that by freezing a pretrained English Transformer language model body and only \textit{lexically} (embedding layers) transferring it to another language, they can later plug in those embeddings into a fine-tuned downstream English model, achieving zero-shot transfer on the downstream task in the other language. \newcite{DBLP:journals/corr/abs-2103-05247} also work with a decoder-only architecture.
% They are specifically interested in the usefulness of encoded natural language knowledge in a pretrained Transformer language model when it comes to fine-tuning towards downstream tasks in other modalities including vision, numerical computation, and protein fold prediction.
They show that by only fine-tuning the input layer, output layer, positional embeddings, and layer norm parameters of an otherwise frozen Transformer language model, they can match the performance of a model fully trained on the downstream task in several modalities.

\paragraph{Lightweight Fine-tuning.} \newcite{houlsby2019parameter} reduce the number of parameters to be updated by inserting adapter modules in every layer of the Transformer model. Then during fine-tuning, they update the adapter parameters from scratch and fine-tune layer norm parameters while keeping the rest of the parameters frozen.
Since adapters are only inserted and initialized at the time of fine-tuning, they are not able to reveal anything about the importance of pretrained modules. Our approach, however, enables highlighting the crucial role of the encoded translation knowledge by contrasting \texttt{\{src,tgt\}+xattn} and \texttt{\{src,tgt\}+randxattn}.
% It is important to distinguish between the adapter modules and cross-attention layers. While adapter modules start from scratch, our \{src,tgt\}+randxattn experiments showed the importance of pretrained values for the cross-attention layers.
\newcite{bapna-firat-2019-simple} devise adapters for MT by inserting language pair-specific adapter parameters in the Transformer architecture. In the multilingual setting, they show that by fine-tuning adapters in a shared pretrained multilingual model, they can compensate for the performance drop of high-resource languages incurred by shared training. \newcite{philip-etal-2020-monolingual} replace language pair-specific adapters with monolingual adapters, which enables adapting under the zero-shot setting.% to pairs never seen before together. 

Another family of lightweight fine-tuning approaches \cite{li-liang-2021-prefix,hambardzumyan-etal-2021-warp,DBLP:journals/corr/abs-2104-08691}, inspired by prompt tuning \cite{NEURIPS2020_1457c0d6}, also relies on updating a set of additional new parameters from scratch towards each downstream task. Such sets of parameters equal a very small fraction of the total parameters in the pretrained model. By contrast, our approach updates a subset of the model's own parameters instead of adding new ones. We leave a comparison of the relative advantages and disadvantages of these approaches to future work.

%termed \textit{prefixes} \cite{li-liang-2021-prefix} or continuous / soft prompts \cite{hambardzumyan-etal-2021-warp, DBLP:journals/corr/abs-2104-08691}

\paragraph{Cross-lingual Embeddings.} Finally, while we were able to obtain cross-lingual embeddings through our transfer learning approach without using any dictionaries or direct parallel corpora, \newcite{DBLP:journals/corr/abs-2010-14649} use a direct parallel corpus and a shared LSTM model that does translation and reconstruction at the same time to obtain aligned embeddings. Given tremendously large monolingual corpora for embedding construction, cross-lingual embeddings can also be obtained by applying a linear transformation on one language's embedding space to map it to the second one in a way that minimizes the distance between equivalents in the shared space according to a dictionary \cite{DBLP:journals/corr/MikolovLS13,xing-etal-2015-normalized, artetxe-etal-2016-learning}. These works specifically targeted the parallel dictionary reconstruction task, while we used the task incidentally, to intrinsically evaluate the parameters learned by our methods.
% Further work has shown that under the assumption that monolingual embedding spaces are isomorphic, fully-supervised guidance from a dictionary is not necessary and cross-lingual mappings can be learned using only a small seed dictionary \cite{artetxe-etal-2017-learning} or with no supervision at all, by relying on matching the distributions of the source and target spaces \cite{artetxe-etal-2018-robust,DBLP:journals/corr/abs-1710-04087}. Such cross-lingual embeddings have been used to initialize models or phrase tables to improve machine translation both in supervised \cite{Kim2019EffectiveCT} and unsupervised settings \cite{DBLP:conf/iclr/ArtetxeLAC18, DBLP:conf/iclr/LampleCDR18, DBLP:conf/emnlp/LampleOCDR18, artetxe-etal-2018-unsupervised,DBLP:conf/acl/ArtetxeLA19,kim-etal-2018-improving,DBLP:conf/acl/PourdamghaniAGK19}.

\section{Conclusion}

We look at how powerful cross-attention can be under constrained transfer learning setups. We empirically show that cross-attention can single-handedly result in comparable performance with fine-tuning the entire Transformer body, and it is through no magic: it relies on translation knowledge in the pretrained values to do so and has new embeddings align with corresponding parent language embeddings. We furthermore show that such aligned embeddings can be used towards catastrophic forgetting mitigation and zero-shot transfer. We hope this investigative study encourages more analyses in the same spirit towards more insights into the inner workings of different modules and how they can be put to good use.

\section*{Acknowledgements}

The authors would like to thank Kushal Chawla, Muhao Chen, Katy Felkner, Thamme Gowda, Xuezhe Ma, Meryem M\textquotesingle{hamdi}, and Xusen Yin for their helpful feedback on the pre-submission draft of this work. This research is based in part on research sponsored by the Office of the Director of National Intelligence (ODNI), Intelligence Advanced Research Projects Activity (IARPA), via AFRL Contract FA8650-17-C-9116 and in part on research sponsored by Air Force Research Laboratory (AFRL) under agreement number FA8750-19-1-1000. Ren's research is supported in part by the Office of the Director of National Intelligence (ODNI), Intelligence Advanced Research Projects Activity (IARPA), via Contract No. 2019-19051600007, the Defense Advanced Research Projects Agency with award W911NF-19-20271, NSF IIS 2048211, and NSF SMA 182926. The views and conclusions contained herein are those of the authors and should not be interpreted as necessarily representing the official policies or endorsements, either expressed or implied, of the ODNI, IARPA, Air Force Laboratory, DARPA, or the U.S. Government. The U.S. Government is authorized to reproduce and distribute reprints for Governmental purposes notwithstanding any copyright annotation thereon.

\bibliography{anthology,custom}
\bibliographystyle{acl_natbib}

\clearpage
% \newpage

% \pagebreak

\appendix

\section{Manual Bilingual Dictionary Evaluation}
\label{sec:appendix}

\begin{table}[ht]
    \centering
    \resizebox{\columnwidth}{!}{\begin{tabular}[width=\columnwidth]{lll}
        \toprule
        German Word & \makecell[l]{\texttt{src+xattn} \\ French Equivalent} & \makecell[l]{\texttt{src+body} \\ French Equivalent} \\
        \hline
        \addlinespace[0.3em]
        Entdeckung & \colorbox{lime}{découverte} & amende \\
        Feind & \colorbox{lime}{ennemi} & \colorbox{lime}{ennemi}\\
        Architekten & \colorbox{lime}{architectes} & architecture \\
        gibt & existe & jette\\
        erforschen & \colorbox{lime}{explorer} & sond \\
        Philosoph & philosophie & philosophie \\
        Cent & centi & centaines \\
        formen & \colorbox{lime}{forme} & \colorbox{lime}{forme} \\
        lassen & \colorbox{lime}{laissez} & PCP \\
        Nummer & \colorbox{lime}{numéro} & Key \\
        können & \colorbox{lime}{puissent} & \colorbox{lime}{puisse} \\
        dasselbe & mêmes & lourds \\
        gelöst & \colorbox{lime}{résoud} & \colorbox{lime}{résoud} \\
        wenig & \colorbox{lime}{peu} & \colorbox{lime}{peu} \\
        zerstört & \colorbox{lime}{détruit} & dévas \\
        Bericht & \colorbox{lime}{reportage} & témoin \\
        Mark & \colorbox{lime}{Mark} & trailer \\
        Brief & \colorbox{lime}{lettre} & \colorbox{lime}{lettres} \\
        Linien & \colorbox{lime}{lignes} & \colorbox{lime}{lignes} \\
        entworfen & \colorbox{lime}{conçus} & monté \\
        Dunkelheit & \colorbox{lime}{ténèbres} & \colorbox{lime}{obscur} \\
        Kreis & \colorbox{lime}{cercle} & \colorbox{lime}{rond} \\
        Haie & \colorbox{lime}{requins} & Hun \\
        spielt & \colorbox{lime}{joue} & tragédie \\
        Elektrizität & \colorbox{lime}{électricité} & électriques \\
        Solar & \colorbox{lime}{solaire} & Arabes \\
        Flügel & \colorbox{lime}{ailes} & avion \\
        Konzept & \colorbox{lime}{concept} & alliance \\
        Strukturen & \colorbox{lime}{structures} & définit \\
        will & \colorbox{lime}{veut} & \colorbox{lime}{voulons} \\
        Hier & \colorbox{lime}{Ici} & Vous \\
        verlieren & \colorbox{lime}{perdent} & \colorbox{lime}{perdent} \\
        unterstützen & \colorbox{lime}{soutien} & \colorbox{lime}{appui} \\
        Planet & \colorbox{lime}{planète} & \colorbox{lime}{planète} \\
        buchstäblich & \colorbox{lime}{littéralement} & multimédia \\
        Schuld & blâ & génére \\
        dass & \colorbox{lime}{que} & toi \\
        plötzlich & \colorbox{lime}{soudainement} & risques \\
        Kann &  \colorbox{lime}{Pouvez} & ciel \\
        Ball & \colorbox{lime}{ballon} & \colorbox{lime}{ballon} \\
        \bottomrule
        \vspace{-7mm} 
    \end{tabular}}
    % \caption{Sampled German words and their equivalents based on the embeddings learned by each of the models. The correct translations are highlighted. Each pair was manually checked for correctness using an automatic translator.}
    % \label{tab:mandict}
\end{table}
% \begin{nolinenumbers}
\noindent Table 4: Sampled German words and their equivalents based on the embeddings learned by each of the models. The correct translations are highlighted. Each pair was manually checked for correctness using an automatic translator.
% \end{nolinenumbers}
% Mozhdeh: I'm sorry about giving in to this solution for this table.

\end{document}